\documentclass[conference]{IEEEtran}
\IEEEoverridecommandlockouts
\usepackage{cite}
\usepackage{amsmath,amssymb,amsfonts}
\usepackage{algorithmic}
\usepackage{graphicx}
\usepackage{textcomp}
\usepackage{xcolor}
\usepackage{multirow}
\usepackage{url}
\usepackage{float}

\usepackage[ruled,vlined]{algorithm2e}
\SetAlFnt{\small}

\definecolor{commentcolor}{RGB}{110,154,155}   % define comment color
\definecolor{keycolor}{rgb}{0.858, 0.188, 0.478} % key color
\newcommand{\PyComment}[1]{\ttfamily\textcolor{commentcolor}{\# #1}}  % add a "#" before the input text "#1"
\newcommand{\PyKey}[1]{\ttfamily\textcolor{keycolor}{#1}} % \ttfamily is the code font

\def\BibTeX{{\rm B\kern-.05em{\sc i\kern-.025em b}\kern-.08em
    T\kern-.1667em\lower.7ex\hbox{E}\kern-.125emX}}
\begin{document}

% Multi-view rObust Representation learnIng via hybRid contrAstive fusioN  
\title{MORI-RAN: Multi-view Robust Representation Learning via Hybrid Contrastive Fusion \\

}

\author{\IEEEauthorblockN{Guanzhou Ke}
\IEEEauthorblockA{\textit{DSIDS \textsuperscript{1}} \\
\textit{Beijing Jiaotong University}\\
Beijing, China \\
guanzhouk@bjtu.edu.cn
\thanks{\textsuperscript{1}Institute of Data Science and Intelligent Decision Support}
}
\and
\IEEEauthorblockN{Yongqi Zhu}
\IEEEauthorblockA{\textit{DSIDS \textsuperscript{1}} \\
\textit{Beijing Jiaotong University}\\
Beijing, China \\
21120632@bjtu.edu.cn}
\and
\IEEEauthorblockN{Yang Yu\textsuperscript{*}}
\IEEEauthorblockA{\textit{DSIDS \textsuperscript{1}} \\
\textit{Beijing Jiaotong University}\\
Beijing, China \\
yangy1@bjtu.edu.cn}
\thanks{\textsuperscript{*} Prof. Yang Yu, the corresponding author, is with the Institute of Data Science and Intelligent Decision Support (DSIDS) and the Frontiers Science Center for Smart High-speed Railway System, Beijing Jiaotong University. }

}

\maketitle

\begin{abstract}
Multi-view representation learning is essential for many multi-view tasks, such as clustering and classification. However, there are two challenging problems plaguing the community: i)how to learn robust multi-view representations from mass unlabeled data and ii) how to balance the view consistency and specificity. To this end, in this paper, we proposed a novel hybrid contrastive fusion method to extract robust view-common representations from unlabeled data. Specifically, we found that introducing an additional representation space and aligning representations on this space enables the model to learn robust view-common representations. At the same time, we designed an asymmetric contrastive strategy to ensure that the model does not obtain trivial solutions. Experimental results demonstrated that the proposed method outperforms 12 competitive multi-view methods on four real-world datasets in terms of clustering and classification. Our source code will be available soon at \url{https://github.com/guanzhou-ke/mori-ran}.
\end{abstract}


\section{Introduction}

% 观点1: 传统的方法无法面对大规模的数据，以及在大规模无标签数据是学习仍然是挑战
% 观点2: 一些办法使用Autoencoder的形式通过重构学习表征，这会导致优化ELBO下界较难，模型需要学习low-level的信息，导致表征不鲁棒，噪音。
%  观点3: 近年来一些对比学习的方法看到了希望，通过对比学习可以使得模型学习概念信息，high-level信息。但是这些方法的任务仍然比较难，如COMPLETE要通过z预测cross-view的z。
% 写作角度：从fusion的形式谈对multi-view representation的重要性，拆分成两个大的类别，显示融合和隐式融合。

\begin{figure}[htbp]
\centerline{\includegraphics[width=0.5\textwidth]{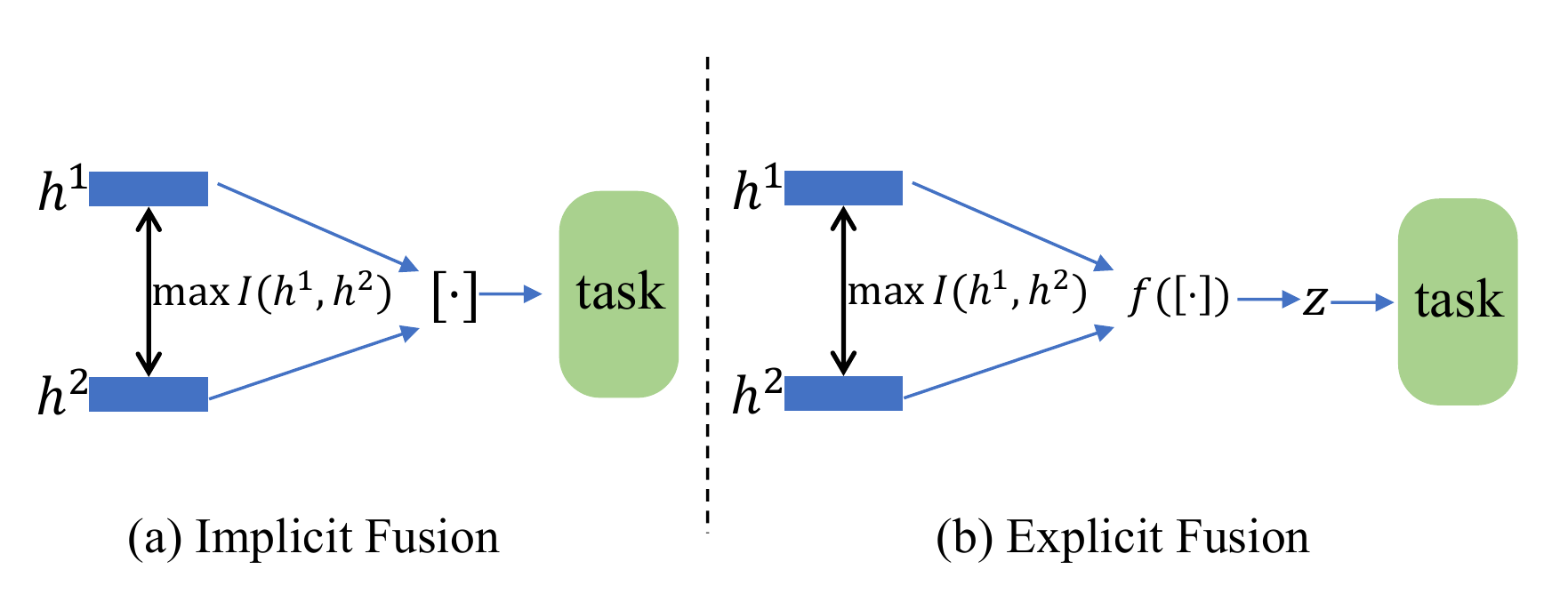}}
\caption{Illustration of the difference between (a) implicit and (b) explicit fusion, where $h$ denotes view-specific representations, $[\cdot]$ is concatenation operation, $f(\cdot)$ implies the fusion function, and $I(\cdot, \cdot)$ is mutual information measurement. Implicit fusion is usually straightforward in maximizing mutual information between view-specific representations, meaning each view representation contains a portion of redundant information. It is then used as input for downstream tasks by concatenation, when the dimensionality of the data to be processed by the downstream task grows linearly with the number of views. Instead, explicit fusion outputs view-specific representations as view-common representations $z$ in either dimension via a mapping function $f(\cdot)$, such as the neural network or the kernel function. Hence that can effectively improve the scalability of downstream tasks.}
\label{fusion-type}
\end{figure}

% 1. 背景，
Multi-view Representation Learning (MRL)\cite{LiYZ19} has received a lot of attention in recent years, especially in downstream tasks such as multi-view clustering and classification \cite{chao2021survey}. In other words, the high-quality representation of multi-view can significantly improve the performance of downstream tasks. Unlike single-view representation learning, the multi-view context requires balancing the representation of heterogeneous information from multiple sources. We argued that methods for processing multi-source information could roughly be divided into two categories: \textit{implicit} fusion \cite{andrew2013deep, wang2015deep, lin2021completer, lin2022dual, sun2022multi, peng2019comic, li_damc} and \textit{explicit} fusion \cite{xue2015gomes, ke2021conan, ke2022efficient, trosten2021reconsidering, xu2021multi, zhou_eamc, SalmanZ20, abavisani2018deep}, as shown in \figurename{\ref{fusion-type}}. Implicit fusion concatenates all view representations as input to a downstream task, while the latter maps all view representations into a more appropriate representation space through a fusion layer, such as the neural network. 

The approach of implicit fusion has received much attention due to the simplicity of the fusion manner. In contrast, explicit fusion requires to design of complex auxiliary tasks to drive the fusion process. However, the former cannot be well suited to address learning from large-scale data and balance view consistency with view specificity \cite{trosten2021reconsidering, ke2021conan}. Some approaches \cite{ke2022efficient, xu2021multi, wang2015deep} have designed fusion modules based on reconstruction to alleviate this. These methods extract common representations from multiple data sources and reconstruct the original data from common representations. Nevertheless, the reconstruction forces the model to learn detailed information about the data, which means that the learned representations contain noise, thus reducing generalization. Although some methods introduce adversarial networks \cite{li_damc, zhou_eamc} to enhance the robustness of multi-view representation, these improvements make the training process unstable.

% 写对比融合
Contrastive learning \cite{hadsell2006dimensionality} has recently emerged as a promising research direction in single-view representation learning \cite{chen2021exploring, he2020momentum, chen2020simclr}. Some new approaches \cite{ke2021conan, trosten2021reconsidering, lin2021completer, lin2022dual} to extend contrastive learning to MRL have also achieved remarkable success. Although these methods can learn high-quality representations from a large amount of unlabeled data, these methods align view-specific representations directly, resulting in a poor balance between view consistency and view specificity. To this end, we propose a novel Multi-view rObust RepresentatIon learnIng method via hybRid contrAstive fusioN (MORI-RAN) to preserve view consistency and specificity. Specifically, we introduce an additional alignment space in the contrastive fusion module to align view-specific representations. At the same time, we leverage instance-level and class-level contrastive objective functions to maximize the mutual information between view-specific and view-common representations. Moreover, we design an asymmetrical contrastive strategy to preserve view diversity information in view-common representations to prevent model collapse. The major contributions of this study are summarized as follows:

\begin{enumerate}
\item We proposed a novel hybrid contrastive fusion method, called MORI-RAN, which can balance the consistency and specificity of views by an additional alignment representation space. Moreover, the ablation study and visualization results demonstrate that the proposed method can preserve the diversity among views without destroying the original representation space.
\item We design an asymmetric representation contrastive strategy to prevent model collapse during the fusion process. The ablation experiment shows that our proposed fusion strategy could learn a more robust representation than the cross-view implicit fusion.

\item We conduct experiments on four real-world datasets to measure the clustering and classification performance of MORI-RAN. Extensive experiments verify the effectiveness of the proposed fusion module.
\end{enumerate}

The remainder of the paper is as follows. Section \ref{related-work} reviews related work, including multi-view representation learning and multi-view fusion. Section \ref{method} introduces the details of the proposed framework, while Section \ref{experiments} presents the experimental results demonstrating the performance of MORI-RAN in different experimental settings. Finally, the discussion and conclusion of our work is presented in section \ref{discussion} and section \ref{conclusion}, respectively. 

\section{Related Work}
\label{related-work}

\subsection{Multi-view Representation Learning}
Multi-view representation learning aims to learn representations from multi-view data that facilitate extracting rich information for downstream tasks, such as clustering or classification. MRL methods can be roughly divided into two categories: alignment and fusion. The alignment approach focuses on capturing the relationships among multiple resource data, and the other aims to fuse the view-specific representation into a single compact (view-common) representation. The CCA-based methods \cite{andrew2013deep, wang2015deep, yan2015deep} are representative of the alignment. These approaches explore the statistical properties between two views, searching for two projections to map the two views onto a low-dimensional view-common representation. Furthermore, some methods based on distance and similarity have been proposed \cite{bronstein2010data, lyu2021multi}. These methods address the cross-view similarity learning problem by embedding multi-view data into a common metric space. Alignment-based approaches have achieved awe-inspiring success for tasks such as information extraction and recommender systems. This paper falls into multi-view fusion, and we will describe related works on fusion in the next part.

\subsection{Multi-view Fusion}
The Multi-view fusion aims to extract view-common representations of arbitrary dimensions from multiple view-specific representations instead of the direct concatenation of all view-specific representations \cite{andrew2013deep, wang2015deep, lin2021completer, lin2022dual, peng2019comic, li_damc}. Although the concatenation strategy may appear more intuitive, it also means that the computational pressure is transferred to the downstream task. Our approach falls into the explicit fusion. Unlike other shallow fusions, such as the kernel fusion\cite{sun2022multi, du2015robust}, the subspaces fusion\cite{zhao2014subspace, abavisani2018deep}, or the weight-sum fusion\cite{zhou_eamc}, we leverage the neural network as the fusion mapper. The benefit of neural networks is the ability to map representations to feature spaces of arbitrary shape. The literature on \cite{zhou_eamc, ke2022efficient} employed the attention mechanism fusion technique. These methods require a long training time and a large amount of training data to converge. Some other methods \cite{ke2021conan, trosten2021reconsidering} using contrastive fusion can alleviate these drawbacks, but literature \cite{trosten2021reconsidering} aligns directly on view-specific representations, which could destroy the diversity between views. The literature \cite{ke2021conan} drives network training based on clustering, which is less compatible for various downstream tasks.

\section{Method}
\label{method}
Our goal is to learn a set of robust view-common representation $z$ from $n$ data points consisting of $V$ views $ \mathcal{D} = \{\textbf{X}^1, \textbf{X}^2, ..., \textbf{X}^V\}$, where $\textbf{X}^v \in \mathbb{R}^{n \times d_{v}}$ denotes the samples of the dimension $d_v$ from the $v$-th views. In this work, we suggest a non-linear multi-view fusion method, i.e., hybrid contrastive fusion, to obtain view-common representations $\bold{z}$. In this section, we first describe the network architecture of MORI-RAN. Next, we introduce the hybrid contrastive fusion method and the asymmetrical contrastive strategy. Finally, we describe the objective function of MORI-RAN.

\subsection{Networks Architecture}\label{section-arch}
\begin{figure*}[!t]
\normalsize
\centerline{\includegraphics[width=1\textwidth]{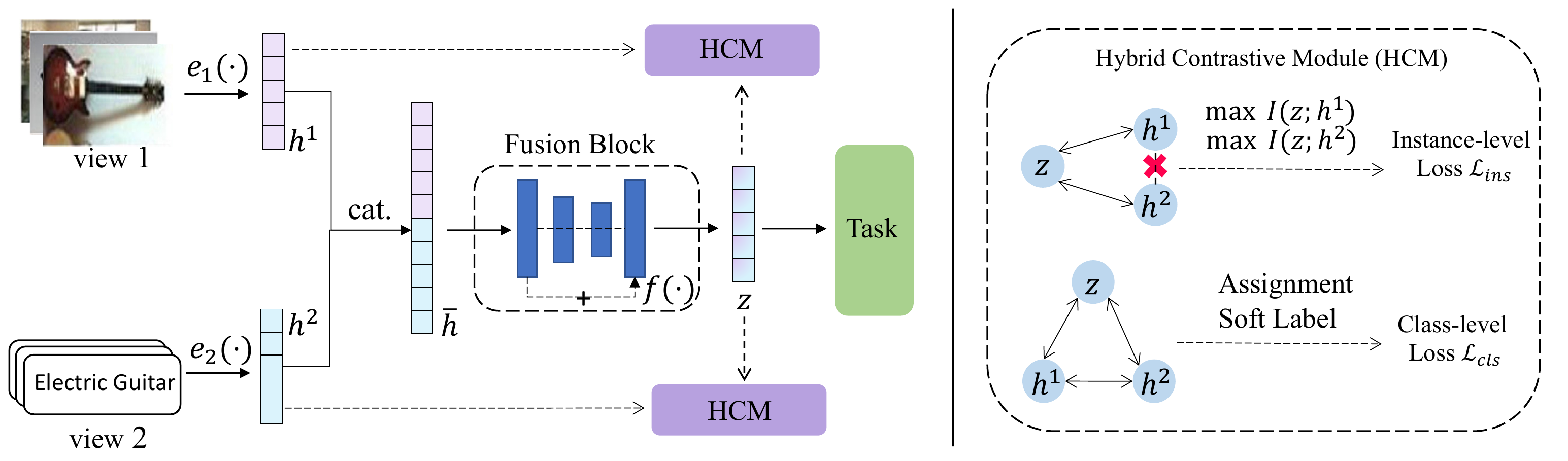}}
\caption{Illustration of MORI-RAN. It (left) consist of $V$ view-specific encoder networks $e_v(\cdot)$ to obtain view-specific representations $\bold{h^v}$, a fusion block $f(\cdot)$ to obtain the view-common representations $\bold{z}$ via fusion the concatenating vectors $\bold{\bar{h}}$. We employed multiple linear layers in the fusion block with ReLU non-linear activation function and leveraged skip-connection to connect the input and output's embeddings. In HCM (right), we treat $\bold{z}$ and $\bold{\bar{h}}$ as input, and we employed an asymmetrical contrastive strategy to maximize the mutual information between $\bold{z}$ and each $\bold{\bar{h}}$. Moreover, we apply the soft label assignment method on $\bold{z}$ and each $\bold{\bar{h}}$, reducing their category-agnostic information.}
\label{framework}
\end{figure*}

The proposed approach, MORI-RAN, consist of $V$ view-specific encoder networks $e_v(\cdot)$, a fusion block $f(\cdot)$, and a hybrid contrastive module (HCM), as illustrated in \figurename{\ref{framework}}.

\begin{itemize}
\item View-specific encoder networks $e_v(\cdot)$ that extracts view-specific representations(vectors) $h$ from each view data point, i.e., $\bold{h^v} = e_v(\bold{X_v})$. MORI-RAN is model-agnostic, meaning that various forms of the network architecture can be treated as encoders, e.g., Fully-Connected Networks (FCNs) or Convolution Neural Networks (CNNs). For simplicity, we employ FCNs to build view-specific encoders.

\item A fusion block $f(\cdot)$ that fusions concatenating vectors $\bold{\bar{h}} = [\bold{h^1}, ..., \bold{h^v}]$ to obtain view-common representations $\bold{z} = f(\bold{\bar{h}})$, where $\bold{\bar{h}} \in \mathbb{R}^{n \times (V \times d_{h})}, \bold{z} \in \mathbb{R}^{n \times d_{h}}$. We build it with fully-connected layers with ReLU and employ the skip connection to connect the fusion block's input and output. The major advantage of our fusion block is that it can map any complex data into a compatible representation space\cite{he2016deep}.

\item The hybrid contrastive module (HCM) treats view-common representations $\bold{\bar{z}}$ and view-specific representations $\bold{\bar{h}}$ as inputs. It leverage instance-level contrastive method to maximize the mutual information between $\bold{\bar{z}}$  and $\bold{\bar{h}}$ with the asymmetrical contrastive strategy, reducing the view-redundant representations with class-level contrastive method. We describe the detail of HCM in the next part.
\end{itemize}

\subsection{The Hybrid Contrastive Module and The Asymmetrical Strategy}\label{section-hc-as}

% 2. Hybrid 的意义
We introduce an additional alignment space $\bold{z}$ in the fusion block. Our motivation is that learns robust view-common representations $\bold{z}$. In other words, we desire to balance the view consistency and specificity. To achieve this, we design the hybrid contrastive module (HCM), in which we leverage the instance-level contrastive method to preserve view-specific information, i.e., we aim to maximize the mutual information between views, as shown in \eqref{eq_maxI}. Moreover, to maximize the view consistency, we optimize the consistency of the assignment of soft labels for view-specific representations $\bold{h}$ and view-common representations $\bold{z}$ using the class-level contrastive method\cite{caron2020unsupervised}. We integrate and trade-off the instance-level and the class-level contrastive method into HCM, elaborating their objective functions in the next part. 
\begin{equation}
\label{eq_maxI}
 \max I(\bold{z}; \bold{h}^v); \quad \text{where} \quad v = 1, 2, \cdots , V
\end{equation}

% 1. 对比学习在多视图背景下带来的弊端，HCM

However, in a multi-view setting, it is prone to model collapse if directly performed contrastive method between view-common representations $\bold{z}$ and view-specific representations $\bold{h}$. It is because $\bold{z} = f([\bold{h}^1, \cdots , \bold{h}^v])$, so a trivial solution is that f equals the constant mapping function, meaning that $\bold{z}$ has not learned effective representations for downstream tasks.

% 3. 非对称如何克服
To this end, we designed an asymmetrical contrastive strategy to prevent the model from obtaining trivial solutions. Concretely, unlike previous methods \cite{lin2021completer, lin2022dual, trosten2021reconsidering}, we disable the alignment between each view-specific representation h and allow only view-specific representations and the introduced view-common representation z to be aligned. The experimental result demonstrates that the model can learn robust representations when employing the asymmetric strategy only on the instance-level contrastive method. 

\subsection{The Objective Function}\label{section-objection-func}

With the above definitions, we design the following objective function for MORI-RAN:

\begin{equation}
\label{total-loss}
\mathcal{L} = \lambda_{1} \mathcal{L}_{ins} + \lambda_{2}\mathcal{L}_{cls}
\end{equation}
where $\mathcal{L}_{ins}$ and $\mathcal{L}_{cls}$ are instance-level contrastive loss and class-level contrastive loss, respectively. The parameters $\lambda_{1}$ and $\lambda_{2}$ are the trade-off factors for instance-level and class-level contrastive loss, respectively. According to our experimental results, we simply fix these two parameters to 1 and 0.5, respectively. We describe the detail of the objective function in the following.

\subsubsection{The instance-level contrastive loss}
In previous works \cite{lin2021completer, lin2022dual, trosten2021reconsidering, ke2021conan}, various forms of contrastive loss have been proposed, such as InfoNCE-based\cite{chen2020simclr}, momentum-based\cite{he2020momentum}, etc. However, all these works require a larger batch size or extra memory during training \cite{tain2020cmc}. These limitations become bottlenecks for multi-view learning. Therefore, inspired by \cite{zbontar2021barlow}, we extend the redundancy-reduction principle to the multi-view. We define the multi-view cross-correlation matrix as following:

\begin{equation}
\label{ccm}
\mathcal{C}_{ij} \triangleq \frac{\sum_b z_{b,i} h^v_{b,j}}{\sqrt{\sum_b (z_{b,i})^2} \sqrt{\sum_b ( h^v_{b,j})^2}} \   \text{where} \ v = 1, 2, \cdots , V
\end{equation}
where $b$ is batch samples and $i, j$ denotes the representation dimension of the networks' outputs. $\mathcal{C}$ is a square matrix of the same size as the dimension of the networks' output, with values range $[-1, 1]$. According to the asymmetric contrastive strategy, we fix the view-common representations $\bold{z}$ and then compute the cross-correlation between $\bold{z}$ and each view-specific representation $\bold{h}$. Finally, we compute the instance contrastive loss as the following:

\begin{equation}
\label{loss-ins}
\mathcal{L}_{ins} = \underbrace{\sum_{i}(1-\mathcal{C}_{ii})^2}_{\text{invariance term}} + \lambda_{ins} \underbrace{\sum_{i}\sum_{j \neq i} \mathcal{C}_{ij}^2}_{\text{redundancy reduction term}}
\end{equation}
where $\lambda_{ins}$ indexes a positive constant trade-off factor. According to the original setting \cite{zbontar2021barlow}, we set $\lambda_{ins} = 5 \times 10^{-3}$ in our experiments.

\subsubsection{The class-level contrastive loss}
We require that the view-common representation $\bold{z}$ and view-specific representations $\bold{h}^v$ be consistent in terms of the soft label assignment. Therefore, we define the class-level contrastive loss as following:
\begin{equation}	
\label{loss-cls}
\mathcal{L}_{cls}(\bold{z}, \bold{h}) = - \sum_{k} Q_{(k)}(\bold{z})\log P_{(k)}(\bold{z}, \bold{h})
\end{equation}
where $k$ denotes the number of soft label, and $p(\cdot, \cdot)$ is defined as the following:

\begin{equation}
	P_{(k)}(\bold{z}, \bold{h}) = \frac{\exp(\bold{z}^Tg_{(k)}(\bold{h})/\tau)}{\sum_{i \neq k}\exp(\bold{z}^Tg_{(i)}(\bold{h})/\tau)}
\end{equation}
where $g(\cdot)$ denotes the soft label assignment network, with output $k$ dimension vector. For simplicity, we employ a MLP to build $g(\cdot)$. $\tau$ is a temperature parameter \cite{wu2018unsupervised}, we set $\tau=0.07$ in our experiments.

Furthermore, we employ the Sinkhorn algorithm \cite{cuturi2013sinkhorn} to compute the optimal soft label assignment of $\bold{z}$, the transportation plan as following:

\begin{equation}
	Q_{(k)}(\bold{z}) := \{ \bold{T} \in \mathbb{R}_{+}^{k \times b} | \bold{T1}_b = \frac{1}{k}\bold{1}_k, \bold{T}^T\bold{1}_k = \frac{1}{b}\bold{1}_b\}
\end{equation}
where $\bold{T}$ is the transportation matrix, and $\bold{1}$ denotes the identity matrix. $b$ is the size of mini-batch. According to the previous work's conclusion \cite{asano2019self}, we can obtain a discrete optimal solution $\bold{T}^*$ by using a rounding procedure. Concretely, the algorithm \ref{alg1} summarizes the proposed method.

\begin{algorithm}[h]
\label{alg1}
\SetAlFnt{\small}
\caption{MORI-RAN pseudocode, PyTorch-style}
\SetAlgoLined
    \PyComment{e$_v$: $v$-th view-specific encoder.} \\
    \PyComment{f: the fusion block.} \\
    \PyComment{g: the soft label assignment network.} \\
    \PyComment{p: projection mlp.} \\
    \PyComment{l\_ins: the instance-level contrastive loss, Eq.4 .} \\
    \PyComment{l\_cls: the class-level contrastive loss, Eq.5.} \\
    \texttt{\\}
    \PyKey{for} x1,...,xv \PyKey{in} dataloader: \\
    \Indp   % start indent
        \PyComment{Step 1: obtain view-specific representations}\\
        h1,...,hv = e1(x1),...,ev(xv) \\
        hs = cat(h1,...,hv) \PyComment{n-by-(v*d)}\\
        \PyComment{Step 2: fusion view-specific representations} \\
        z = f(hs) \\
        \PyComment{Step 3: compute the instance-level loss} \\
        L += INS(z, hs) \\
        \PyComment{Step 4: compute the class-level loss} \\
        L += CLS(z, hs) \\
        L.backward() \\
        update(e, f, g, p) \\
    \Indm % end indent, must end with this, else all the below text will be indented
    \texttt{\\}
    \PyKey{def} INS(z, hs) \PyComment{instance-level contrastive loss} \\
    \Indp
        hs = split(hs) \PyComment{to list.} \\ 
        zp = p(z) \PyComment{projection.} \\
        hps = [p(h) \PyKey{for} h \PyKey{in} hs] \\
        subloss = 0. \\
        \PyComment{compute zp and each hp loss .} \\
        \PyKey{for} hp \PyKey{in} hps: \\
            \Indp
            subloss += l\_ins(zp, hp) \\
            \Indm
    \PyKey{return} subloss \\
    \Indm
    
    \texttt{\\}
    \PyKey{def} CLS(z, hs) \PyComment{class-level contrastive loss} \\
    \Indp
        all = cat(z, hs) \PyComment{to list.} \\ 
        allc = [g(h) \PyKey{for} h \PyKey{in} hs] \\
        subloss = 0. \\
        \PyComment{Traverse the `allc` and get the pairwise element} \\
        \PyKey{for} c1, c2 \PyKey{in} pairwise(allc): \\
            \Indp
            subloss += l\_cls(c1, c2) \\
            \Indm
    \PyKey{return} subloss \\
    \Indm
    \texttt{\\}
\end{algorithm}

\section{Experiments}
\label{experiments}
Several experiments are conducted to study and evaluate the performance of MORI-RAN on four real-world datasets. Specifically, we evaluate MORI-RAN on the clustering and the classification task. 

\subsection{Dataset}
We evaluate MORI-RAN and other competitive methods using four well-known multi-view datasets containing image and text data. There are:
\begin{itemize}
\item NoisyMNIST \cite{yang2010bag}, which is a large-scale and widely-used benchmark dataset consisting of 70k handwritten digit images with $28 \times 28$ pixels. We use the multi-view version provided by \cite{wang2015deep}, which consists of the original digits and the white Gaussian noise version, respectively. As most of the comparison method cannot process such a large dataset, we only use a subset of NoisyMNIST consisting of 10k validation images and 10k testing images.
\item Scene-15 \cite{fei2005bayesian}, which contains 4,485 images with outdoor and indoor scene environments. We use the multi-view version provided by \cite{lin2021completer}, which consists of PHOG and GIST features.
\item LandUse-21 \cite{yang2010bag} consists of 2100 satellite images from 21 categories with PHOG and LBP features.
\item Caltech101-20 \cite{fei2004learning}. We use the multi-view version provided by \cite{li2015large}, which consists of 2,386 images of 20 subjects with the views of HOG and GIST features.
\end{itemize}
We report the dataset description in Table \ref{dataset_description}.

\begin{table}[htbp]
\renewcommand{\arraystretch}{1.3}
\caption{Dataset Description}
\label{dataset_description}       % Give a unique label
\begin{center}
\begin{tabular}{ccccc}
\hline\noalign{\smallskip}
Dataset & \#samples & \#view & \#class & \#shape \\
\noalign{\smallskip}\hline\noalign{\smallskip}
NoisyMNIST & 20,000 & 2 & 10 & (784, 784) \\
Scene-15 & 4,485 & 2 & 15 &  (20, 59) \\
LandUse-21 & 2,100 & 2 & 21 & (59, 40) \\
Caltech101-20  & 2,386 & 2 & 20 & (1984, 512) \\
\noalign{\smallskip}\hline
\end{tabular}
\end{center}
\end{table}

\subsection{Baseline Models}

We compare the proposed method with 12 multi-view representation learning baseline methods, including: 1) Deep Canonical Correlation Analysis (DCCA) \cite{andrew2013deep}, 2) Deep Canonically Correlated Auto-encoders (DCCAE) \cite{wang2015deep}, 3) Binary Multi-view Clustering (BMVC) \cite{zhang2018binary}, 4) Efficient and Effective Regularized Incomplete Multi-view Clustering (EERIMVC) \cite{liu2020efficient}, 5) Partial Multi-View Clustering (PVC) \cite{li2019reciprocal}, 6) Autoencoder in autoencoder networks (Ae$^2$-Nets) \cite{zhang2019ae2}, 7) Doubly Aligned Incomplete Multi-view Clustering (DAIMC) \cite{hu2019doubly}, 8)Incomplete MultiModal Visual Data Grouping (IMG) \cite{zhao2016incomplete}, 9)Unified Embedding Alignment Framework (UEAF) \cite{wen2019unified}, 10)Efficient Multi-view Clustering Networks (EMC-Nets) \cite{ke2022efficient}, 11) Contrastive fusion newtorks for multi-view clustering (CONAN) \cite{ke2021conan}, and 12) Incomplete Multi-view Clustering via Contrastive Prediction (COMPLETER) \cite{lin2021completer}. Briefly, for DCCA and DCCAE, we fix the dimensionality of the hidden representation to 10. For BMVC, we fix the length of the binary code to 128. For EERIMVC, we construct the kernel matrix using a "Gaussian kernel" and use the optimal $\lambda$ in the interval from $2^{-15}$ to $2^{15}$ with an interval of $2^3$ \cite{lin2021completer}.

\subsection{Implementation Details}
We implement MORI-RAN and other non-linear comparison methods on the PyTorch 1.10 \cite{paszke2019pytorch} platform, running on Ubuntu 18.04 LTS utilizing a NVIDIA A100 tensor core Graphics Processing Units (GPUs) with 40 GB memory size. We describe the implementation details in the following:
\subsubsection{Networks setting}
For simplicity, we set the dimensionality of view-specific encoders to $I-1024-1024-1024-O$ for all experiments, where $I$ and $O$ indicate the dimension of the data's input and the encoder's output, respectively. Concretely, we set $O=128$ for Scene-15, LandUse-21, and Caltech101-20, while set $O=288$ for NoisyMNIST. For the fusion block, we set the dimensionality of network to $O-O/2-O$, and leverage the skip connection to connect the input and output vector. Moreover, we set the dimensionality of the projection network to $O-O\times4-O$.
\subsubsection{Training}
We use the Adam optimization technique \cite{KingmaB14} with default parameters and an initial learning rate of $0.0001$ for both clustering and classification. For clustering, we treat the whole dataset as the training set and validation set, shuffling the training set during training. We train the model for 100 epochs and set the size of the mini-batch to 256 for all experiments. Specifically, we train the model 5 times with the different random seeds and report the result from the lowest value of Eq. \eqref{total-loss}. For classification, we split the dataset into a training set, a validation set, and a test set in the ratio $8:1:1$. We first train the network for 90 epochs without labels. Then, we leverage the method provied by\cite{lin2022dual} to generate the predicted label. Finally, we use the test set to evaluate the classification performance of MORI-RAN.

\subsection{Evaluation Metrics}
For clustering, the performance is measured using three standard evaluation matrices, i.e., Accuracy (ACC), Normalized Mutual Information (NMI), and Adjusted Rand Index (ARI). For further details on these evaluation metrics, the reader is referred to \cite{kumar2011co}. It should be noted that the validation process of the clustering methods involves only the cases where ground truth labels are available. For classification, ACC, Precision, and the F-Score are used. For all measurements, a higher value indicates better performance.

\subsection{Compared with Baseline Models}

\begin{table*}
\renewcommand{\arraystretch}{1.3}
\caption{Clustering results on four benchmark datasets. The best and the second best values are highlighted in {\color{red}red} and {\color{blue}blue}, respectively.}
\label{clustering_result}       % Give a unique label
\begin{center}
\begin{tabular}{ccccccccccccc}
\hline\noalign{\smallskip}
 & \multicolumn{3}{c}{NoisyMNIST} & \multicolumn{3}{c}{Scene-15} & \multicolumn{3}{c}{LandUse-21} & \multicolumn{3}{c}{Caltech101-20} \\
\noalign{\smallskip} Method & ACC & NMI & ARI & ACC & NMI & ARI & ACC & NMI & ARI & ACC & NMI & ARI \\
\noalign{\smallskip}\hline\noalign{\smallskip}
DCCA (2013) \cite{andrew2013deep} & 85.53 & 89.44 & 81.87  & 36.18 & 38.92 & 20.87 & 15.51 & 23.15 & 4.43 & 41.89 & 59.14 & 33.39 \\
DCCAE (2015) \cite{wang2015deep} & 81.60 & 84.69 & 70.87  & 36.44 & 39.78 & 21.47  & 15.62 & 24.41 & 4.42 & 44.05 & 59.12 & 34.56 \\
BMVC (2019) \cite{zhang2018binary} & 81.27 & 76.12 & 71.55  & 40.50 & 41.20 & 24.11  & 25.34 & 28.56 & 11.39 & 42.55 & 63.63 & 32.33  \\
EERIMVC (2020) \cite{liu2020efficient} &  65.47 & 57.69 & 49.54  & 39.60 & 38.99 & 22.06  & 24.92 & 29.57 & 12.24 & 43.28 & 55.04 & 30.42 \\
PVC (2019)  \cite{li2019reciprocal} & 41.94 & 33.90 & 22.93  & 30.83 & 31.05 & 14.98  & 25.22 & 30.45 & 11.72 & 44.91 &  62.13 & 35.77 \\
Ae$^2$-Nets (2019) \cite{zhang2019ae2}  & 56.98 & 46.83 & 36.98  & 36.10 & 40.39 & 22.08 & 24.79 & 30.36 & 10.35 & 49.10 & 65.38 & 35.66 \\
DAIMC (2019) \cite{hu2019doubly} & 39.18 & 35.69 & 23.65  & 32.09 & 33.55 & 17.42 &  24.35 & 29.35 & 10.26 & 45.48 & 61.79 & 32.40\\
IMG (2016) \cite{zhao2016incomplete} & - & - & -  &  24.20 & 25.64 & 9.57 &  16.40 & 27.11 & 5.10 & 44.51 & 61.35 & 35.74\\
UEAF (2019) \cite{wen2019unified} &  67.33 & 65.37 & 55.81  & 34.37 & 36.69 & 18.52 & 23.00 & 27.05 & 8.79 & 47.40 & 57.90 & 38.98\\
EMC-Nets (2022) \cite{ke2022efficient} & 68.71 & 70.30 & 69.17  & 31.05 & 29.87 & 16.97 & 14.81 & 25.67 & 7.66 & 43.17 & 37.11 & 29.64\\
CONAN (2021) \cite{ke2021conan} & 39.33 & 33.62 & 22.39  & 36.77 & 37.33 & 20.33 & \color{blue}26.70 & 29.30 & 11.47 & 29.09 & 34.09 & 18.53\\
COMPLETER (2022) \cite{lin2021completer} & \color{blue}\color{blue}89.08 & \color{blue}88.86 & \color{blue}85.47  &  \color{blue}41.07 & \color{red} 44.68 & \color{blue}24.78 &  25.63 & \color{blue}31.73 & \color{blue}13.05 & \color{red} 68.91 & \color{blue}65.36 & \color{red} 75.17\\
MORI-RAN (Ours) & \color{red}95.22 & \color{red}89.28 & \color{red}89.83  & \color{red}43.80 & \color{blue}43.81 & \color{red}25.44 & \color{red}28.60  & \color{red}33.95 & \color{red}14.22 & \color{blue}65.73 & \color{red}67.19 & \color{blue}54.61 \\
\noalign{\smallskip}\hline
\end{tabular}
\end{center}
\end{table*}

\begin{table*}
\renewcommand{\arraystretch}{1.3}
\caption{Classification results on four benchmark datasets.}
\label{classification_result}       % Give a unique label
\begin{center}
\begin{tabular}{ccccccccccccc}
\hline\noalign{\smallskip}
 &  \multicolumn{3}{c}{NosiyMNIST} & \multicolumn{3}{c}{Scene-15} & \multicolumn{3}{c}{LandUse-21} & \multicolumn{3}{c}{Caltech101-20} \\
\noalign{\smallskip} Method & ACC & P(recision) & F-Score & ACC & P & F-Score & ACC & P & F-Score & ACC & P & F-Score \\
\noalign{\smallskip}\hline\noalign{\smallskip}
BMVC (2019) \cite{zhang2018binary} & 90.17 & 89.40 & 87.33 & 42.72 & 40.65 & 34.57  & 43.57 & 45.92 & 41.01  & 76.61 & 73.17 & 68.29  \\
EERIMVC (2020) \cite{liu2020efficient} & 81.20 & 80.98 & 81.07 & 53.71 & 54.69 & 47.45  & 37.47 & 35.49 & 33.90  & 53.59 & 48.09 & 49.44   \\
EMC-Nets (2022) \cite{ke2022efficient} & 73.19 & 71.10 & 70.84 & 45.24 & 38.69 & 39.47  & 27.66 & 25.71 & 26.03 & 50.18 & 48.68 & 49.31 \\
CONAN (2021) \cite{ke2021conan} & 55.13 & 52.34 & 49.29 & 41.15 & 40.31 & 37.10  & 30.82 & 26.11 & 25.44 & 29.71 & 24.03 & 23.51 \\
COMPLETER (2022) \cite{lin2021completer} & \color{blue}93.17 & \color{blue}90.62 &  \color{blue}90.30 & \color{blue}75.10 & \color{blue}74.00  & \color{blue}73.40 & \color{blue}69.67 & \color{red}70.80 & \color{red}70.40 & \color{red}91.93 & \color{blue}84.20 & \color{blue}83.80 \\
MORI-RAN & \color{red}97.53 & \color{red}93.15 & \color{red}90.76  & \color{red}78.80 & \color{red}\color{red}75.84 & \color{red}75.13 & \color{red}71.74  & \color{blue}70.21 & \color{blue}69.83 & \color{blue}88.35 & \color{red}85.61 & \color{red}84.57 \\
\noalign{\smallskip}\hline
\end{tabular}
\end{center}
\end{table*}

We evaluate our model and baseline models on four well-known datasets and report the performance results on clustering and classification in Table \ref{clustering_result} and Table \ref{classification_result}, respectively. For clustering, the results show that MORI-RAN can have a significant improvement compared to baseline models. It is worth noting that the proposed method manages a performance improvement of $6.14\% (95.22 - 89.08)$, $0.42\%  (89.28 - 88.86)$, and $4.35\%  (89.83 - 85.47)$ against the second-best method on the NosiyMNIST dataset in terms of ACC, NMI, and ARI metrics, respectively. Moreover, our method significantly improves over other explicit fusion methods, such as CONAN\cite{ke2021conan} and EMC-Nets\cite{ke2022efficient}, on all four datasets. Nevertheless, our method shows a significant superiority over CONAN\cite{ke2021conan}, which also uses the contrastive fusion method. For classification, the experimental results show that our proposed method performs the classification task better on the four datasets. We also validate the effectiveness of the explicit fusion when compared with the second best method, COMPLETER \cite{lin2021completer}.

\subsection{Ablation Study}

\begin{table}
\renewcommand{\arraystretch}{1.3}
\caption{The Ablation Study of different modules on the NosiyMNIST dataset.}
\label{ab_module}       % Give a unique label
\begin{center}
\begin{tabular}{cccccc}
\hline\noalign{\smallskip}
$\mathcal{L}_{ins}$ & $\mathcal{L}_{cls}$ & Asym. & ACC & NMI & ARI \\
\noalign{\smallskip}\hline\noalign{\smallskip} $\checkmark$ & $\checkmark$ & $\checkmark$ & 95.22 & 89.28 & 89.83 \\
 $\checkmark$ & -- & -- & 88.12\color{gray}(-7.09) & 85.64\color{gray}(-3.64) & 85.64\color{gray}(-4.18) \\
 $\checkmark$ & -- & $\checkmark$ & 90.47\color{gray}(-4.75) & 87.33\color{gray}(-1.95) & 85.70\color{gray}(-4.12) \\
 -- & $\checkmark$  & -- & 36.10\color{gray}(-59.12) & 35.92\color{gray}(-53.36) & 17.53\color{gray}(-72.3) \\
 -- & $\checkmark$  & $\checkmark$ & 32.48\color{gray}(-62.74) & 28.90\color{gray}(-60.38) & 10.14\color{gray}(-79.69) \\
  $\checkmark$ & $\checkmark$  & -- & 91.00\color{gray}(-4.21) & 86.59\color{gray}(-2.68) & 86.59\color{gray}(-3.23) \\
\noalign{\smallskip}\hline
\end{tabular}
\end{center}
\end{table}

We conducted a series of ablation studies on the NoisyMNIST dataset for each component of MORI-RAN, i.e., the instance-level contrastive module ($\mathcal{L}_{ins}$), the class-level contrastive module ($\mathcal{L}_{cls}$), and asymmetric fusion strategy (Asym.), and the results are shown in Table \ref{ab_module}. The experimental results show that the instance-level contrastive module significantly affects the quality of view-common representation $\bold{z}$, and the opposite of the class-level contrastive module degrades the generalization performance of downstream tasks when used alone. Furthermore, we found that the asymmetrical contrastive strategy is effective for the instance-level contrastive module, but the performance is the lowest among all combinations when it is used in combination with the class-level contrastive module. Also, we noted that the clustering performance is lower than the baseline of MORI-RAN when the instance-level contrastive loss without the asymmetrical strategy. Therefore, we believe that the hybrid contrastive fusion with asymmetrical contrastive strategy is capable of learning robust representations.

\subsection{Visualization}

\begin{figure*}[!t]
\normalsize
\centerline{\includegraphics[width=1\textwidth]{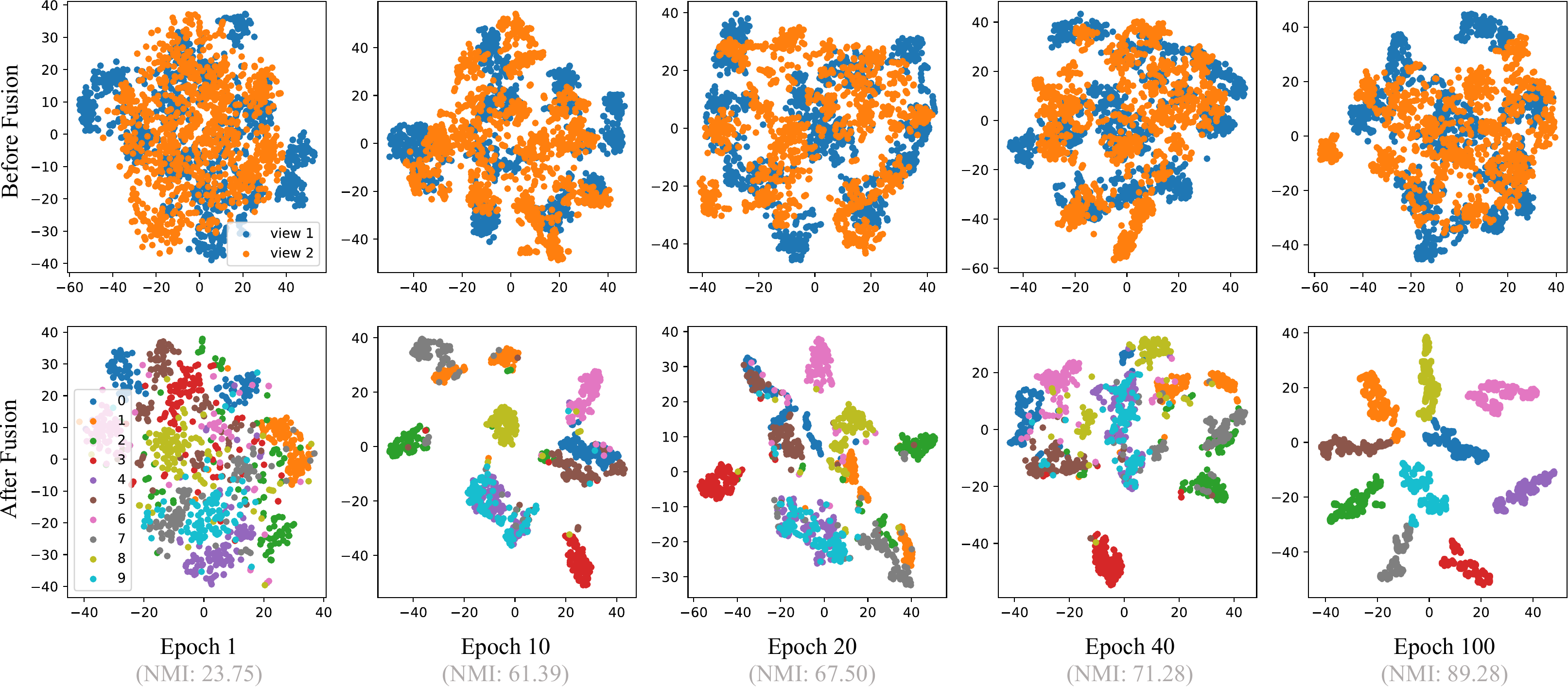}}
\caption{Visualization of representations on the NosiyMNIST dataset before (first row) and after fusion (second row) using T-SNE \cite{van2008visualizing} at epoch 1st, 10th, 20th, 40th, and 100th.}
\label{visualization}
\end{figure*}

We demonstrate the change of representations before and after contrastive fusion on the NosiyMNIST dataset, as shown in \figurename{\ref{visualization}. For simplicity, we randomly choose 1K samples projected to 2-D using T-SNE \cite{van2008visualizing}. The visualization demonstrated that the view-specific representations $\bold{h}^1$ and $\bold{h}^2$ preserve their self-structure during training. At the same time, the common representation $\bold{z}$ becomes more separable and compact with increasing the training epoch. Therefore, we argue that the HCM can improve the quality of view-common representation and does not harm the view-specific representations. 

\section{Discussion}
\label{discussion}

Although our method achieves superior performance on four datasets, it does not make a significant margin with the second-best method on the image-text dataset. Therefore, we argue that MORI-RAN has some limitations in the multimodal downstream tasks. In this study, we have used some smaller or medium-scale datasets. However, more large-scale datasets have been released \cite{wang2022chiqa, gu2022wukong}, and we will try to extend MORI-RAN to fit super-large datasets in the future. It is worth noting that MORI-RAN is not a reconstruction-based approach to learning multi-view representations, and we claim that reconstruction is a challenging task. However, we have noticed some promising works \cite{he2022masked, wei2022masked} emerging recently, and we will try to balance the contrastive approach and the reconstruction approach in future work.

\section{Conclusion}
\label{conclusion}
We proposed a novel hybrid contrastive fusion method and an asymmetrical contrastive strategy for learning robust multi-view representation. This study introduced an additional alignment space to preserve the multi-view diversity and extract the multi-view shared information. The ablation studies demonstrated that the instance-level contrastive module with an asymmetrical strategy could preserve the multi-view diversity. The class-level contrastive module could make the view-common structure more separable and compact. Experiments on four datasets show that the proposed method performs the clustering and classification tasks well compared to the other 12 competitive methods.

\section*{Acknowledgment}
This work was supported by the Fundamental Research Funds for the Central Universities (Science and technology leading talent team project), China (No. 2021JBZD006), and the Fundamental Research Funds for the Beijing Jiaotong University, China (No. 2021JBWZB002).

\bibliographystyle{IEEEtran}
\bibliography{paper.bib}

\end{document}